\documentclass[letterpaper,10pt,journal,twoside]{IEEEtran}
\usepackage[english]{babel}
\addto\captionsenglish{\renewcommand{\figurename}{Fig.}\renewcommand{\tablename}{TABLE}}
\usepackage[T1]{fontenc}

\usepackage[nolist]{acronym}
\usepackage[cmex10]{amsmath}
\usepackage{amssymb}
\usepackage{bm}
\usepackage{booktabs}
\usepackage{color}
\usepackage{graphicx}
\usepackage{hyperref}
\usepackage{import}
\usepackage{mathtools}
\usepackage{multirow}
\usepackage{multicol}
\usepackage{flushend}
\usepackage{outline}
\usepackage{paralist}
\usepackage{siunitx}
\usepackage{stmaryrd}
\usepackage{systeme}
\usepackage{url}
\usepackage{tikz}
\usepackage{booktabs}
\usepackage{balance}
\usepackage{placeins}
\usepackage{needspace}
\usepackage{cite}
\usetikzlibrary{tikzmark}
\usepackage{esvect}
\usepackage[normalem]{ulem}
\usepackage{bbm}
\usepackage{siunitx}
\usepackage{wrapfig,lipsum,booktabs}

\usepackage{graphicx}
\usepackage{adjustbox}
\usepackage{xcolor}
\usepackage{flushend}
\usepackage{booktabs}
\usepackage{makecell }

\usepackage{bbding}  
\usepackage{pifont}  

\usepackage{tabularx}
\usepackage{array}

\newcommand{\rev}[1]{{\ifdefined\ShowChanges\color{blue}\else\color{black}\fi#1}}
\newcommand{\changedgraphic}[1]{\ifdefined\ShowChanges\begingroup\setlength{\fboxsep}{-0.4pt}\setlength{\fboxrule}{0.4pt}\fcolorbox{blue}{white}{#1}\endgroup\else#1\fi}
\newenvironment{revision}{\begingroup\ifdefined\ShowChanges\color{blue}\else\color{black}\fi}{\endgroup}
\newcommand{\vnew}[1]{\leavevmode{\ifdefined\ShowChanges\color{blue}\else\ifdefined\ShowIncrementalChanges\color{blue}\else\color{black}\fi\fi#1}}
\newsavebox{\revisiongraphicbox}
\renewcommand{\changedgraphic}[1]{\begingroup\leavevmode\sbox{\revisiongraphicbox}{#1}\usebox{\revisiongraphicbox}\ifdefined\ShowChanges\llap{\smash{\color{blue}\setlength{\fboxsep}{-0.4pt}\setlength{\fboxrule}{0.4pt}\fbox{\phantom{\usebox{\revisiongraphicbox}}}}}\fi\endgroup}

\usepackage{colortbl}
\definecolor{ourcolor}{HTML}{99e0eb}
\definecolor{ourblue}{HTML}{27a2c3}
\definecolor{tablecolor}{HTML}{ccf2f5} 
\definecolor{tablecolor2}{HTML}{ffcdb4}
\definecolor{citecolor}{HTML}{fe7b5b}
\definecolor{grey}{rgb}{0.9, 0.9, 0.9}

\definecolor{gred}{rgb}{0.859,0.267,0.216}
\definecolor{ggreen}{rgb}{0.059,0.616,0.345}
\definecolor{deepblue}{HTML}{27a2c3}
\definecolor{deepred}{HTML}{fe7b5b}

\newcommand{\ours}{\textit{PAKE} }

\newcommand{\ddpm}[2]{$#1\scriptstyle{\pm#2}$}
\newcommand{\ddpmbf}[2]{\cellcolor{tablecolor}$\mathbf{#1\scriptstyle{\pm#2}}$}

\newcommand{\cc}[1]{$#1$}
\newcommand{\ccbf}[1]{\cellcolor{tablecolor}$\mathbf{#1}$}

\title{PAKE: Learning Whole-Body Loco-Manipulation with Partial Kinematic Embeddings}
\author{Zhengmao~He$^{1*}$, Moonkyu~Jung$^{2*}$, Hyeongjun~Kim$^{2}$,\\
Jiseong~Lee$^{2}$, Hui~Zhang$^{3}$, Jemin~Hwangbo$^{2}$, and Jie~Song$^{1,3,4}$%
\thanks{Manuscript received June 29, 2026; revised September 13, 2026;
accepted October 8, 2026.}%
\thanks{This paper was recommended for publication by Editor Sehoon Ha upon
evaluation of the Associate Editor and Reviewers' comments.
\textit{(Corresponding authors: Jemin Hwangbo and Jie Song.)}}%
\thanks{$^{*}$Equal contribution.}%
\thanks{\raggedright $^{1}$The Hong Kong University of Science and Technology (Guangzhou).
Emails: \texttt{zhe037@connect.hkust-gz.edu.cn}, \texttt{jsongroas@hkust-gz.edu.cn}.}%
\thanks{\raggedright $^{2}$Korea Advanced Institute of Science and Technology.
Email: \texttt{moonk127}, \texttt{kaist0914}, \texttt{ssuni1002016}, \texttt{jhwangbo@kaist.ac.kr}.}%
\thanks{\raggedright $^{3}$ETH Z\"urich.}%
\thanks{\raggedright $^{4}$The Hong Kong University of Science and Technology.}%
}
\hypersetup{hidelinks,
pdftitle={PAKE: Learning Whole-Body Loco-Manipulation with Partial Kinematic Embeddings},
pdfauthor={Zhengmao He, Moonkyu Jung, Hyeongjun Kim, Jiseong Lee, Hui Zhang, Jemin Hwangbo, Jie Song}}
\AtBeginDocument{\raggedbottom}

\begin{document}
\bstctlcite{PAKE:IEEEreferenceStyle}

\maketitle

\begin{abstract}
Loco-manipulation has recently shown promising capabilities; however, achieving high-precision control, managing the high-dimensional action space induced by many degrees of freedom (DoFs), and fully exploiting the inherent redundancy of whole-body systems remain challenging.
In this paper, we propose a novel whole-body control framework that effectively addresses these challenges by decomposing the complex loco-manipulation problem into partial reference motion generation and low-level imitation control. We introduce a new Kinematic Normalizing Flow (KNF) model, trained on a large-scale kinematic dataset, that generates diverse partial kinematic references. A high-level controller is then trained to navigate the KNF's latent space to exploit redundant solutions, while a low-level controller \rev{learns dynamically consistent motion execution}. We validate our approach on the quadrupedal robot equipped with a six-DoF robotic arm. In simulation, PAKE outperforms the evaluated baselines in tracking accuracy and feasible workspace coverage on the 4,000-target benchmark. For hardware deployment, we evaluate the system over 24 episodes across 8 different mobile loco-manipulation tasks. The system achieves end-effector pose-tracking errors of 4.5 cm and 0.14 rad, while maintaining accurate locomotion tracking with linear and angular velocity errors of 0.1 m/s and 0.01 rad/s, respectively.
Our method represents a practical and powerful solution for accurate and generalized whole-body loco-manipulation in high-DoF robotic systems, with promising potential for diverse downstream robotic tasks.

\end{abstract}
\IEEEpeerreviewmaketitle

\section{Introduction}



\IEEEPARstart{M}obile manipulation has garnered significant attention in recent years, primarily because it greatly expands the workspace available to robotic manipulators. Manipulators mounted on legged chassis have shown unique advantages due to their enhanced capability to overcome obstacles. A floating-base platform, in particular, introduces additional degrees of freedom (DoF), allowing robotic arms to significantly extend their workspace and perform whole-body control tasks. Furthermore, by fully utilizing the redundancy provided by these extra DoFs, the robotic arm can assist with locomotion tasks even while tracking target objects.





Recently, model-free Reinforcement Learning (RL) has become increasingly popular in the fields of both locomotion \cite{hwangbo2019learning, choi2023deformable} and manipulation \cite{zhang2025RobustDexGrasp, huang2024fungrasp}, as it effectively facilitates control policy learning through extensive interaction within simulated environments. However, loco-manipulation poses several challenges that differentiate it from simpler fixed-base manipulation tasks or pure locomotion scenarios. These challenges make it difficult for RL approaches to achieve high-precision solutions or fully leverage the redundant solutions inherent in high-DoF systems.

\begin{figure}[t]
    \centering
    \includegraphics[width=\linewidth, trim={0 0 40mm 0}, clip]{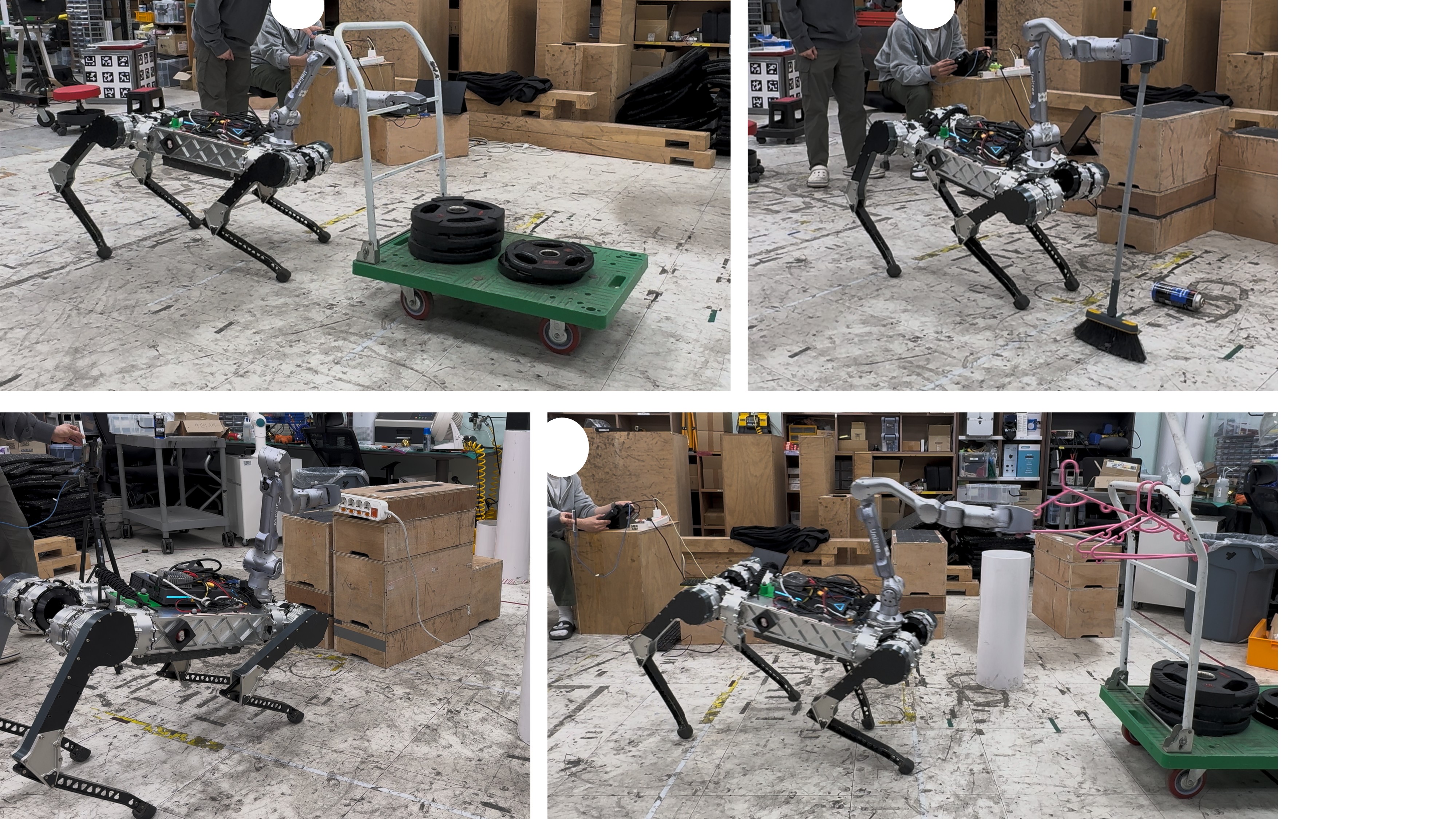}
    \caption{
    Real-world deployments of the proposed framework across diverse tasks, including pulling a heavy cart, sweeping the floor, plugging in a charger, and hanging a clothes hanger.
    }
    \label{fig:teaser}
\end{figure}

These challenges include:

(\romannumeral1) \textit{Search space:} High degrees of freedom create an extremely large search space, complicating the reinforcement learning process of identifying meaningful and coordinated actions for both manipulation and locomotion. Consequently, many existing methods are often evaluated or applied in scenarios where the robot remains stationary during manipulation~\cite{portela2024wholebodyendeffectorposetracking, fu2022deep}.

(\romannumeral2) \textit{Coordination complexity:} Locomotion and manipulation involve distinct and sometimes conflicting objectives. If coordination is not managed properly, movements intended for one purpose may adversely impact the other. For instance, even minor vibrations in the robot's torso can be greatly magnified at the end-effector, compromising precision tasks. Similarly, shifts in the robot’s center of mass (CoM) and momentum due to manipulator motion can negatively affect locomotion stability.



To address these challenges, numerous approaches have been proposed; however, most existing works~\cite{fu2022deep, liu2024vbc, portela2023learningforce, he2024legmanip, arm2024pedipulate} have primarily focused on position tracking. Although some studies~\cite{pan2024roboduetwholebodyleggedlocomanipulation, fu2022deep} attempted to track both position and orientation simultaneously, their methods resulted in significant errors. Other research~\cite{portela2024wholebodyendeffectorposetracking, jiang2025learning} has achieved higher-precision pose tracking, but at the cost of chassis mobility by transforming the robot into a fixed-base manipulator or by adding extra wheels to minimize leg movement, thereby simplifying exploration.
More importantly, existing approaches often tackle the complex, high-dimensional action space by searching for just a single feasible solution. These approaches overlook the system's inherent redundancy—a vast space of alternative solutions that could be exploited to address complex coordination challenges and enhance tracking precision.

In this work, we propose a hierarchical framework for whole-body loco-manipulation that decouples redundancy-aware kinematic reference generation from dynamically feasible execution. We first learn a Kinematic Normalizing Flow (KNF) from a kinematic dataset. The learned KNF provides a compact latent embedding of the robot’s kinematic redundancy, mapping a desired end-effector pose to a distribution of candidate partial reference motions. These partial references specify only upper-body configurations, while leaving the remaining full-body DoFs, especially the leg DoFs required for locomotion, to be recovered by the controller. 
The  high-level controller (HLC) outputs actions in the KNF latent space, which are interpreted as latent variables conditioned on the desired end-effector pose. These latent variables are then decoded by the KNF into partial reference motions, such as torso and arm reference trajectories. In this way, the HLC selects among redundant kinematic solutions encoded by the KNF, allowing it to generate references that facilitate end-effector tracking and can be effectively realized by the LLC.
Since these KNF-generated references are obtained from kinematic data alone, they may violate physical constraints such as joint limits, leg reachability, or locomotion stability. To bridge this gap, the low-level controller (LLC) \rev{tracks the torso references through leg-joint control, while the arm tracks KNF joint references through PD control}. In this way, the HLC exploits the redundancy encoded by the KNF, while the LLC \rev{learns to adapt reference execution to the robot’s dynamics and maintain stable whole-body motions}.

In summary, our main contributions are:

(\romannumeral1) A decoupled control framework that simplifies the exploration for loco-manipulation by separating high-level, redundancy-aware kinematic planning and low-level dynamic execution.

(\romannumeral2) A redundancy-aware partial-kinematic representation, implemented with KNF, that provides a structured high-level action space for coordinating locomotion and manipulation.

(\romannumeral3) Extensive sim-to-real validation that demonstrates state-of-the-art tracking performance and showcases the robustness and applicability of our method.

\label{sec:Introduction}

\section{Related Work}

\subsection{Locomotion}

Research on robotic locomotion has significantly matured in recent years. Early efforts primarily focused on model-based approaches, which utilized simplified robot models to enable real-time computation using Model Predictive Control (MPC). These methods addressed issues such as model inconsistencies arising from contact and off-ground states by defining explicit contact sequences~\cite{bell2016perception,Carlo2018cheetah,kim2019wbc,Ruben2023nonlinear}. However, such approaches often lack flexibility and struggle to manage unexpected contacts or slippage. More recently, reinforcement learning (RL) methods have demonstrated substantial potential by training controllers to achieve specific locomotion tasks through extensive interaction data collected in simulation environments~\cite{raisim, makoviychuk2021isaac}, subsequently enabling effective transfer to real robots via sim-to-real techniques. RL approaches have successfully accomplished tasks such as velocity tracking~\cite{hwangbo2019learning}, fall recovery~\cite{hwangbo2019learning, ma2023fall}, gait diversity learning~\cite{yang2020multiexpert, margolis2022walkwaystuningrobot}, and parkour-style movements in complex terrains combined with visual perception~\cite{hoeller2023anymal,cheng2023parkour,zhuang2023robot}, significantly improving the robustness of controllers. Nonetheless, using reinforcement learning to achieve precise end-effector target tracking for high-DoF locomotion systems remains an open challenge.

\subsection{Whole-body control}

In recent years, significant research has focused on loco-manipulation, particularly emphasizing whole-body control (WBC). Model-based approaches allow robots to leverage their full degrees of freedom (DoF) for precise target tracking by accurately modeling the robotic system dynamics~\cite{sleiman2021mpc, belliscoso2019alma, ferrolho2023roloma}. However, these methods depend heavily on accurate system modeling and predefined contact sequences, leading to limited flexibility.
Recently, reinforcement learning (RL)-based methods have demonstrated considerable promise for high-DoF legged robotic systems~\cite{hwangbo2019learning, lee2020learning}. For instance, Ma et al.\cite{ma2022combining} combined RL-based locomotion control with model-based arm manipulation; however, the decoupled nature of their approach restricts whole-body coordination. Fu et al.\cite{fu2022deep} tackled whole-body control through end-to-end RL with a mixed-advantage and curriculum learning strategy but achieved only position tracking with the robot stationary. Liu et al.\cite{liu2024vbc} proposed a hierarchical approach, combining RL-based leg control with an inverse Jacobian controller for arm motion; nevertheless, joint limitations caused local minima issues, significantly constraining the arm’s operational workspace. Portela et al.\cite{portela2023learningforce} developed a loco-manipulation approach for force control, but their method exhibited limited position-tracking accuracy. Ha et al.\cite{ha2024umilegs} implemented task-specific pose tracking relative to the world frame but lacked independent chassis movement control. Portela et al.\cite{portela2024wholebodyendeffectorposetracking} achieved high-precision 6D pose tracking, although their solution was restricted to fixed-base configurations and relied heavily on supplementary locomotion controllers. More recently, Jung et al.\cite{jung2026Learningdynamic} proposed a hierarchical RL framework enabling quadrupedal mobile manipulators to perform adaptive, dynamic pick-and-place with heavier payloads and extended workspaces.


\rev{PAKE explicitly represents and learns to select from redundant partial-IK solutions.} Given the redundancy available in such high-DoF systems, multiple valid solutions exist for achieving a specific 6D pose. 
Utilizing these redundant solutions effectively can improve the motion performance of robot, as well as enable more sophisticated downstream tasks such as collision avoidance.


\section{Method}

\begin{figure*}[t]
    \centering
    \includegraphics[width=0.88\linewidth, trim={25bp 65bp 105bp 4bp}, clip]{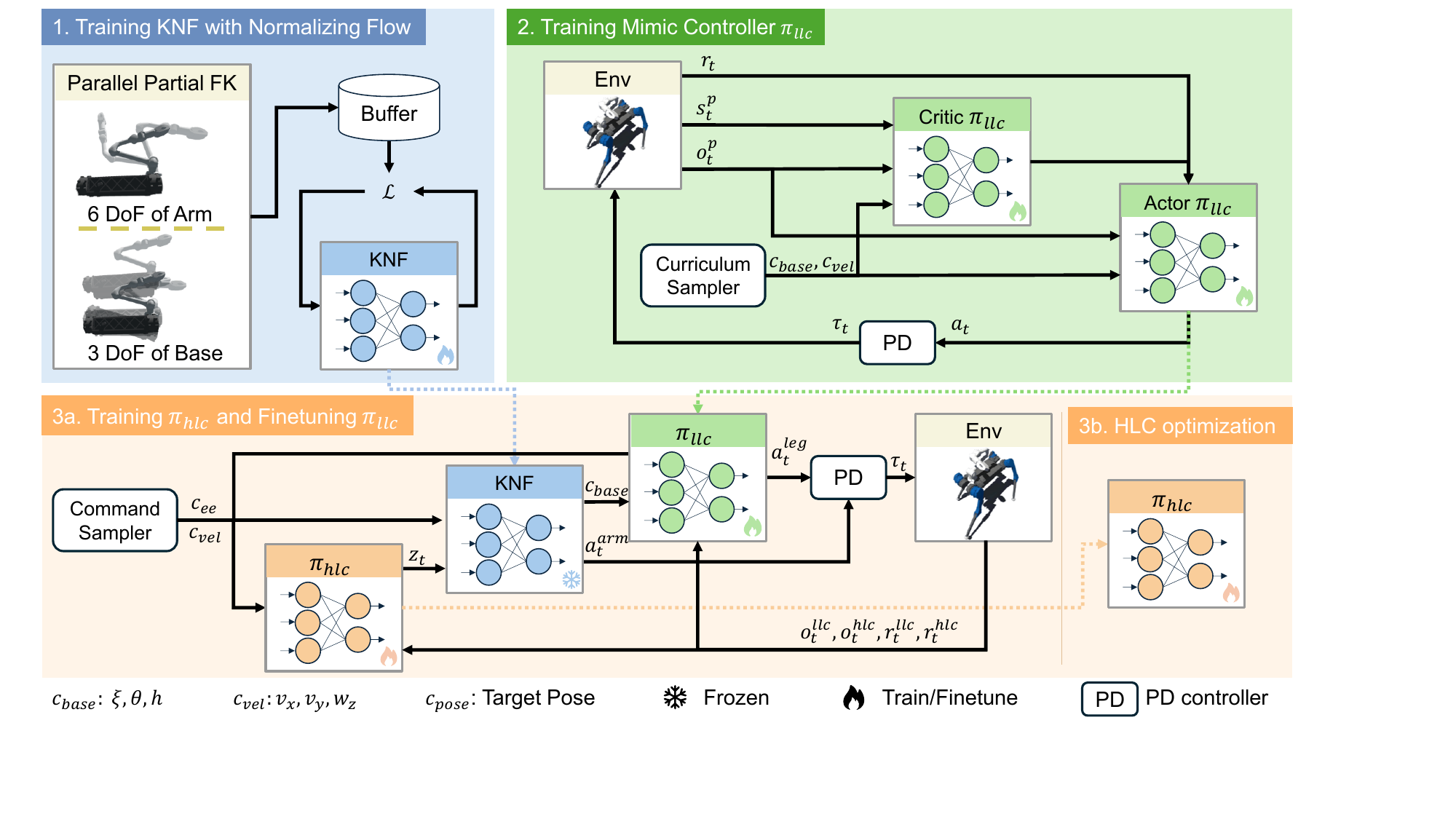}
    \caption{
    (1) \vnew{Train} \vnew{KNF} \vnew{on} \vnew{partial} \vnew{torso}\vnew{--}arm kinematics. (2) \vnew{Pretrain} \vnew{the} \vnew{LLC} to track torso references \vnew{and} \vnew{locomotion} \vnew{commands}. (3) \vnew{Train} the \vnew{HLC} \vnew{to} \vnew{select} \vnew{KNF} latent \vnew{inputs} \vnew{while} \vnew{fine-tuning} the \vnew{LLC}. \rev{Joint training is followed by teacher-guided HLC-only optimization.} KNF \rev{generates} \rev{torso} \rev{and} \vnew{arm-joint} \vnew{references}\vnew{;} \vnew{the} \vnew{LLC} \vnew{controls} \vnew{the} \vnew{legs}\vnew{,} \vnew{and} \rev{arm} \vnew{PD} \vnew{control} \vnew{tracks} \vnew{KNF} \rev{joint} \vnew{targets}\rev{.}
}
    \label{fig:framework}
\end{figure*}

\subsection{Overview}

Figure~\ref{fig:framework} \vnew{summarizes} \vnew{the} \vnew{three-stage} \vnew{PAKE} \vnew{training}. First, we \vnew{sample} \vnew{6} \vnew{arm} \vnew{and} \vnew{3} \vnew{torso} \vnew{DoFs} using parallel forward kinematics \vnew{and} \vnew{learn} \vnew{a} \vnew{kinematic} \vnew{embedding} \vnew{with} \vnew{Normalizing} \vnew{Flows}\vnew{~}\vnew{\cite{ames2022ikflow}}. \vnew{End-effector} poses are \vnew{expressed} \vnew{in} the heading frame~\cite{fu2022deep,liu2024vbc}, defined by the \vnew{robot's} \vnew{$xy$}\vnew{-}\vnew{position} and yaw (Figure~\ref{fig:sketch}). \vnew{The} embedding \vnew{forms} the \vnew{HLC} action space.

Next, we pretrain \vnew{the} \vnew{LLC} to track \rev{velocity} \rev{commands} \rev{and} \rev{torso} \vnew{references} \rev{through} \rev{leg-joint} \rev{control}\rev{.} \vnew{KNF} \vnew{references} are partial and may exceed leg \vnew{reachability} \vnew{or} joint limits, or compromise walking stability.

Finally, \vnew{the} \vnew{HLC} \vnew{learns} to \vnew{select} \vnew{redundant} KNF \rev{solutions}. Concurrently, we fine-tune the low-level imitation policy to follow dynamically changing torso reference trajectories, \rev{learning} \rev{to} \rev{adapt} \rev{their} \rev{execution} \rev{to} the \vnew{robot's} \rev{dynamics} \rev{while} \rev{maintaining} \rev{locomotion}\rev{.} \rev{This} \rev{physics-based} \rev{imitation} \rev{bridges} \rev{the} \rev{gap} \rev{between} kinematic \rev{reference} \rev{generation} and \rev{dynamic} \rev{execution}\rev{.} \rev{Joint training achieves accurate tracking but can leave abrupt references. We seek smoother reference selection while preserving this tracking capability, without permanently restricting the available latent range. Stage~3 therefore continues with teacher-guided HLC optimization, using distillation before further PPO.}

\subsection{Partial Kinematic Embedding}
\subsubsection{Partial Kinematic Dataset}
\label{subsection:dataset}
\begin{figure}[tbp]
    \centering
    \begin{minipage}[c]{0.49\linewidth}
    \includegraphics[width=\linewidth, trim={0 20mm 0 50mm}, clip]{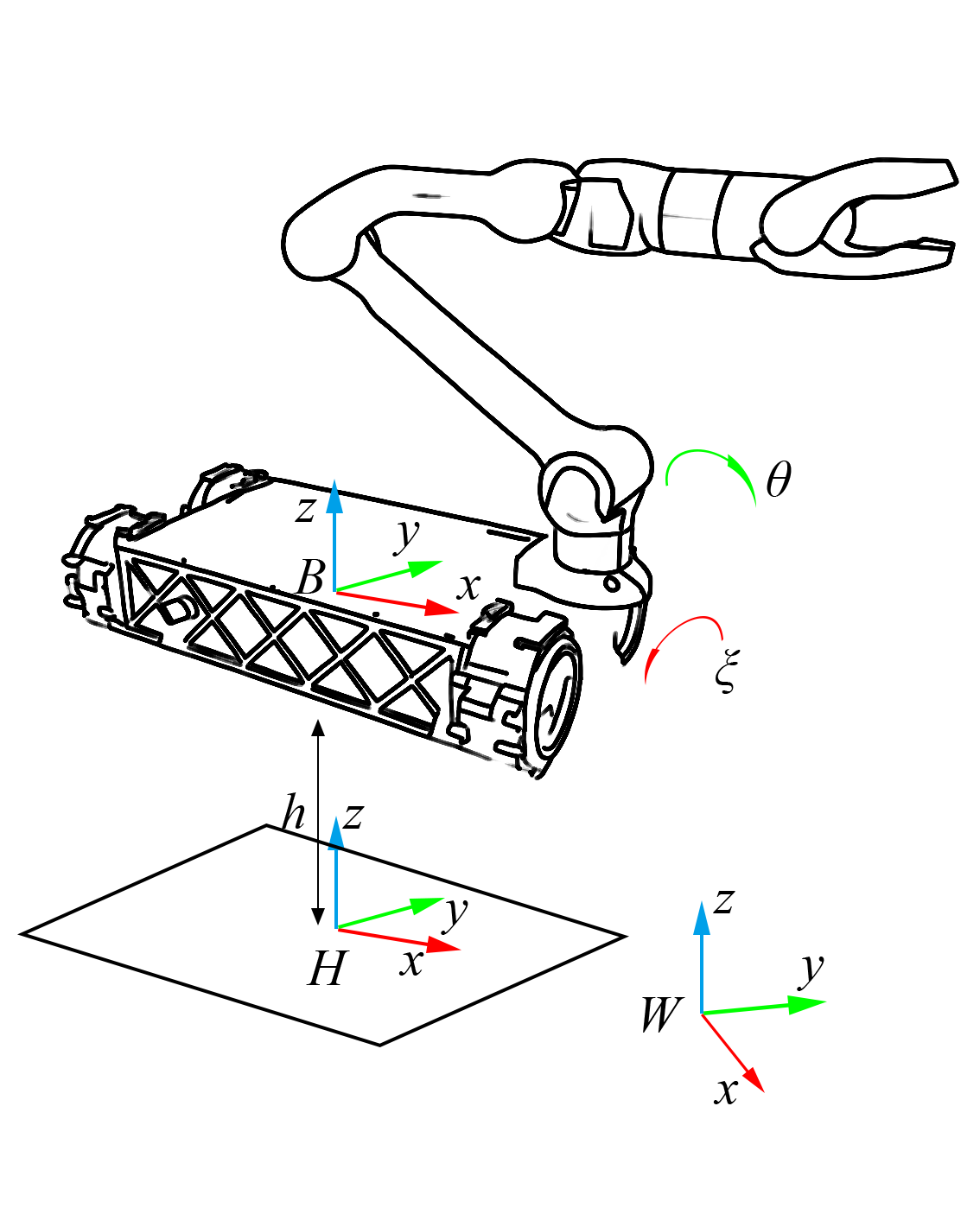}
    \end{minipage}\hfill
    \begin{minipage}[c]{0.49\linewidth}
    \centering\footnotesize
    \begin{revision}
    Sampling ranges\\[2pt]
    \setlength{\tabcolsep}{3pt}
    \begin{tabular}{lrr}
    \toprule Coordinate & Min. & Max.\\\midrule
    $h$ [m] & 0.15 & 0.65\\
    $\xi$ [rad] & $-0.8$ & 0.8\\
    $\theta$ [rad] & $-0.3$ & 0.3\\
    $q_1$ [rad] & $-2.618$ & 2.618\\
    $q_2$ [rad] & 0 & 2.967\\
    $q_3$ [rad] & $-2.880$ & 0\\
    $q_4$ [rad] & $-1.518$ & 1.518\\
    $q_5$ [rad] & $-1.344$ & 1.344\\
    $q_6$ [rad] & $-2.793$ & 2.793\\\bottomrule
    \end{tabular}
    \end{revision}
    \end{minipage}
    \caption{
    \vnew{Partial} \vnew{torso}\vnew{--}\vnew{arm} reference model \vnew{and} \vnew{sampling} \vnew{ranges}. \vnew{Reference} \vnew{configurations} and end-effector \vnew{poses} are expressed \vnew{in} the heading frame.
}
    \label{fig:sketch}
\end{figure}

\vnew{The} partial reference $\mathbf{q}^{i}$ \vnew{comprises} \vnew{torso} roll $\xi$, pitch $\theta$, height $h$, and \vnew{six} arm \vnew{joints} (Figure~\ref{fig:sketch})\vnew{;} \vnew{torso} $x$, $y$, and yaw \vnew{are} \vnew{excluded} to \vnew{preserve} mobility. \vnew{References}\vnew{,} \vnew{end-effector} \vnew{poses} \vnew{$\chi_e$}\vnew{,} \vnew{and} \vnew{3D} \vnew{velocity} \vnew{commands} \vnew{$v_{vel}$} \vnew{use} \vnew{the} \vnew{heading} \vnew{frame}\vnew{.}

\vnew{The} dataset $\mathcal{M}={\mathbf{q}^{i}, \chi_{e}^{i}}$ \vnew{contains} 170 million pairs, \vnew{generated} within seconds \vnew{by} uniformly \vnew{sampling} configurations within their predefined joint limits and \vnew{computing} \vnew{poses} \vnew{with} parallel forward kinematics\vnew{,} $\chi_e=\chi_e(\mathbf{q})$\vnew{,} \vnew{rather} \vnew{than} \vnew{inverse} \vnew{kinematics}. \rev{Leg} \rev{constraints} \rev{are} \rev{omitted} \rev{to} \rev{avoid} \rev{tying} \rev{the} \rev{representation} \rev{to} \rev{a} \rev{fixed} \rev{stance}\rev{;} \rev{contact-dependent} \rev{execution} \rev{is} \rev{learned} \rev{by} \rev{the} \rev{low-level} \rev{controller}\rev{.}

\subsubsection{Learning Kinematic Motion Embedding}
Given the kinematic dataset $\mathcal{M}$, we train a Normalizing Flow (NF) model to generate a distribution of \rev{partial kinematic solutions} conditioned on the target end-effector pose.

NF \vnew{represents} multimodal \vnew{conditional} distributions \vnew{through} \vnew{invertible} \vnew{mappings} between a simple base distribution \vnew{and} the target distribution. \vnew{Sampling} the base distribution \vnew{generates} \vnew{candidate} \vnew{solutions} \vnew{efficiently}.

Our proposed KNF model utilizes the framework introduced by Ames et al.\cite{ames2022ikflow} as its backbone, incorporating additional stabilizing techniques from Kim et al.\cite{kim2020softflow}. This integration enables KNF to efficiently represent most of the solution space with minimal reconstruction error.  \rev{We apply this construction to the partial torso--arm chain and use its latent coordinates as the high-level action space.}

The KNF adopts a normal distribution as its base distribution, simplifying both the computation of the maximum-likelihood loss function and the action space selection for the high-level controller.
We implement coupling layers following the Glow architecture proposed by Kingma and Dhariwal~\cite{kingma2018glow}, employing three-layer fully connected neural networks with a hidden width of 1024 neurons as coefficient functions. These coupling layers encode both input variables and conditional information. Our KNF comprises 12 coupling layers, each with a dimensional width of 12, which exceeds the dimension of the solution space, facilitating the effective modeling of multimodal target distributions. 

\subsection{Pretraining Low-level Imitation Policy}


In this section, we describe the training process for the low-level controller policy $\pi_{llc}$, enabling the robot to track 3D torso reference motions and velocity commands. We formulate the partial motion imitation problem as a Markov Decision Process (MDP), training the robot to follow arbitrary 3D body velocity commands $c_{vel} = (v_{x}, v_{y}, w_{z})$ as well as 3D torso commands $c_{base}$, including desired torso roll $\xi$, pitch $\theta$, and height $h$.

\subsubsection{Network Architecture}

We employ the Proximal Policy Optimization (PPO) algorithm to train the low-level imitation policy, using an asymmetric actor-critic architecture similar to the one proposed by Brakel et al.~\cite{Brakel2022asymmetric}. Both actor and critic networks utilize fully connected layers, with sizes $[256 \times 128]$ for the actor and $[512 \times 256 \times 128]$ for the critic. The actor maps proprioceptive observations to action outputs, while the critic estimates value functions based on the current state.

The actor network outputs action values representing the offset from nominal joint positions for 12\rev{ leg} actuators, which are then added to the nominal positions to obtain target joint positions. These \rev{leg-joint} \rev{targets} are tracked using a proportional-derivative (PD) controller with gains set as $k_p=100$ and $k_d=1$. The actor runs at 100 Hz both in simulation and on the real robot, whereas the PD controller operates at 500 Hz in simulation and 4000 Hz on real hardware. Nominal leg joint positions are defined by the robot's stance during initialization, while nominal arm positions are sampled from a small normal distribution.

\subsubsection{Curriculum-Based Command Sampler}\label{subsection:command}

The commands $c_{vel}$ and $c_{base}$ are sampled at the beginning of each episode and remain fixed throughout. Desired torso roll $\xi$ and pitch $\theta$ are sampled uniformly within intervals centered at zero, with ranges increasing progressively during training iterations, up to a maximum interval of $(-0.26~\text{rad}, +0.26~\text{rad})$. The torso height $h$ is similarly sampled uniformly within a range whose lower bound gradually decreases as training progresses, ultimately spanning from $0.3~\text{m}$ to $0.5~\text{m}$. The velocity command $c_{vel}$ components are sampled uniformly from intervals centered at zero, with their ranges progressively expanding over iterations to maximum values: $v_{x}\in(-1.0~\text{m/s}, 2.0~\text{m/s})$, $v_{y}\in(-1.0~\text{m/s}, 1.0~\text{m/s})$, and $w_{z}\in(-1.0~\text{rad/s}, 1.0~\text{rad/s})$.

\subsubsection{Policy Input}\label{subsection:observation}

The actor network observes proprioceptive states $o^{p}$, which include a 10-step history of leg, arm, and torso observations, a 3-step history of previous controller actions, a 10-step history of end-effector targets, and a 9-dimensional partial reference motion. Gaussian noise is added to these proprioceptive inputs to simulate realistic sensor data.

The critic receives privileged state information $s^{p}$, which consists of noise-free proprioceptive states $o^{p}$, along with additional privileged information obtained directly from simulation. This privileged state includes histories of end-effector states and foot-contact states (each over 10 steps), linear torso velocity, torso height, roll, pitch, as well as foot-end time-of-flight and stance durations.

\subsubsection{Reward}\label{subsubsection:reward}

The reward encourages command tracking, end-effector pose tracking, stable contact behavior, and smooth whole-body motion. It follows common locomotion RL designs~\cite{ji2022concurrent}, with additional terms for end-effector pose tracking, torso posture tracking, and arm-motion regularization. To avoid penalty terms dominating the task rewards, we compute $r=r_{pos}\times \text{exp}(0.1\times r_{neg})$
where most penalties are scaled by curriculum factors. The command curriculum for $c_{\mathrm{vel}}$ and $c_{\mathrm{base}}$ is described in Section~\ref{subsection:command}. The compact reward specification is given in Table~\ref{tab:reward_compact}.

\begin{table}[t]
\scriptsize
\centering
\caption{Compact reward terms.}
\label{tab:reward_compact}
\setlength{\tabcolsep}{2.2pt}
\renewcommand{\arraystretch}{1.08}
\begin{tabularx}{\columnwidth}{@{}p{0.3\columnwidth}Xc@{}}
\toprule
Term & Expression 
\\
\midrule
Velocity tracking
& $\exp\!\left(-\|\mathbf{v}_{xy}-\mathbf{v}^{d}_{xy}\|^2
-1.5\|\omega_z-\omega_z^d\|^2\right)$
\\

EE pose tracking
& $\|\mathbf{p}_{pos}-\mathbf{p}^{d}_{pos}\|^2+
\|\mathbf{p}_{ori,6D}-\mathbf{p}^{d}_{ori,6D}\|^2$
\\
\midrule

Airtime
& $\sum_i r_{\mathrm{air},i}(T_{s,i},T_{a,i})$
\\

Foot contact
& $\sum_i \lambda_c\|(\mathbf{R}_{c,i}\mathbf{v}_{f,i})_{xy}\|^2$
\\

Foot clearance
& $\sum_i (p_{f,i,z}-h_{\mathrm{tar}})^2
\|(\mathbf{R}_{b}^{T}\mathbf{v}_{f,i})_{xy}\|$
\\

Torso posture
& $s\!\left[(h-c_{b,0})^2+(\xi-c_{b,1})^2+(\theta-c_{b,2})^2\right]$
\\

Nominal pose
& $\dfrac{s\lambda_c}{B+1}
\left(\|\mathbf{q}_{leg}-\mathbf{q}_{leg}^{0}\|
+\|\mathbf{q}_{arm}-\mathbf{q}_{arm}^{0}\|\right)$
\\
\midrule

Base regularization
& $v_z^2+0.02|\omega_x|+0.02|\omega_y|$
\\



Leg/Arm regularization
& $\|\boldsymbol{\tau}\|^2,\;
\lambda_c\|\dot{\mathbf{q}}\|^2,\;
\lambda_c\|\rev{\dot{\mathbf{q}}}-\dot{\mathbf{q}}^{prev}\|^2,\;
\lambda_c S(\mathbf{a})$
\\

\bottomrule
\end{tabularx}
\footnotesize{\textbf{Note:}
$h_{\mathrm{tar}}=0.10$.
$s=10$ when standing and $1$ otherwise.
$B=10\sqrt{(0.5-c_{b,0})^2+c_{b,2}^2}$.
$S(\mathbf{a})=0.5\|\mathbf{a}_{t-3}+\mathbf{a}_{t-1}-2\mathbf{a}_{t-2}\|^2
+\|\mathbf{a}_{t-1}-\mathbf{a}_{t-2}\|^2$.
Leg-wise terms are summed over four legs.
}
\end{table}

\subsubsection{Domain Randomization}\label{subsubsection:domainrandomization}
To facilitate effective zero-shot sim-to-real transfer, we apply domain randomization by varying key simulation parameters. Specifically, ground friction, joint friction, and observation noise parameters are randomized at the beginning of each training episode, sampling values from normal distributions within predefined ranges.

\subsection{Training High-level Controller Policy}

In this section, we introduce the third training phase, which combines the trained KNF model with the training of a high-level controller $\pi_{hlc}$. The high-level controller is trained to explore the latent space of the KNF, \rev{learning} \rev{to select from} the \rev{redundant} \rev{partial-IK solutions represented} by the KNF model, while simultaneously fine-tuning the low-level imitation controller $\pi_{llc}$. The goal of this combined approach is to achieve accurate, generalized 6D end-effector pose tracking for high-degree-of-freedom systems, even during highly dynamic motions.

\subsubsection{Implementation Details}

The \vnew{HLC} \vnew{uses} the same asymmetric actor-critic \vnew{network} \vnew{structures} as the \vnew{LLC}, with \vnew{similar} observation and reward designs\vnew{.}

\paragraph{Action Space}
The high-level controller explores the latent space of the KNF, generating latent variables $z$ from its actions through the following transformation: 
\begin{equation}\label{equation:action}
z_{t}=\lambda_{\text{latent\_scale}}\times \text{tanh}(a^{hlc}_{t})
\end{equation}
Here, the hyperparameter $\lambda_{latent\_scale}$ is \rev{set to 0.3}, balancing the diversity and quality of the generated solutions~\cite{ames2022ikflow}.

\paragraph{Rewards}
In addition to the rewards defined in Section~\ref{subsubsection:reward}, the high-level controller also receives a smoothness reward based on the difference between partial reference motions at consecutive timesteps. This encourages the generation of smooth reference trajectories.

\subsubsection{Command Sampling}

\paragraph{End-Effector Trajectory Parameters Sampling}
Unlike most previous works~\cite{fu2022deep, liu2024vbc, ha2024umilegs} that interpolate sampled points to define manipulation trajectories, we directly sample the end-effector target trajectory using third-order rational Bézier curves~\cite{ji2023Hierarchical, he2024legmanip}. These curves effectively represent complex trajectories, including varying velocities and accelerations, more closely resembling real-world manipulation tasks.

The parameters defining each rational Bézier curve include control point positions and their corresponding weights. The control point positions are uniformly sampled within the kinematic range defined by forward kinematics (as described in Section~\ref{subsection:dataset}), while the control point weights are uniformly sampled within the range [1, 2000]. The orientation trajectory is derived through spherical linear interpolation (slerp) between the orientations of the initial and final control points.

\paragraph{Base Command Sampling}

The torso velocity commands are uniformly sampled from the maximum ranges defined in Section~\ref{subsection:command}, with all curriculum scaling factors initialized at their maximum value (set to 1). However, the torso reference motions $c_{base}$ are no longer sampled independently; instead, these motions are generated directly by the KNF model.

\begin{revision}
After 1000 joint HLC--LLC updates at $\lambda_{\mathrm{latent\_scale}}=0.7$, we freeze the KNF, LLC, and normalization. We optimize the HLC for 100 PPO updates at scale 0.035 to obtain a teacher, distill its behavior for 75 updates at scale 0.3, then perform 50 PPO updates with teacher retention (Table~\ref{tab:refinement}). Auxiliary objectives fit context-specific teacher offsets, encourage latent and arm-reference diversity, and penalize temporal changes and excessive reference rates. PAKE-S omits this additional HLC optimization, retaining scale 0.7 and the same inference architecture.
\end{revision}
\begin{table}[t]
\centering\footnotesize
\caption{\rev{HLC optimization following joint training.}}
\label{tab:refinement}
\begin{revision}
\begin{tabular}{lccc}
\toprule Optimization & Updates & $\lambda_{\mathrm{latent\_scale}}$ & LR\\\midrule
PPO & 100 & 0.035 & $5\!\times\!10^{-5}$\\
Distillation & 75 & 0.3 & $5\!\times\!10^{-4}$\\
PPO + teacher retention & 50 & 0.3 & $10^{-5}$\\\bottomrule
\end{tabular}
\end{revision}
\end{table}

\section{Experiments}

In this \rev{section}, we present a series of experiments designed to address the following key research questions:

\begin{itemize} 
    \item \textbf{Q1}: Can \ours enable highly accurate, generalized end-effector tracking while simultaneously maintaining precise chassis locomotion? 
    \item \textbf{Q2}: How significantly does \ours improve whole-body tracking accuracy compared to previous methods? 
    \item \textbf{Q3}: How do HLC training and execution-time latent selection affect tracking performance?
\end{itemize}


\subsection{Implementation Detail}

To efficiently generate large-scale, high-precision data closely resembling real-world interactions, we use RaiSim~\cite{raisim} as our simulation environment and implement our algorithms with PyTorch~\cite{pytorch}. Our experiments are conducted using RAIBO2~\cite{hwangbo2025raibo2} from RAION
Robotics, integrated with a Unitree Z1 manipulator, providing 18 actuators in total with joint position encoders. 

Training was performed using an AMD Ryzen 9 7950X CPU and an NVIDIA GeForce RTX 4080 GPU. Pretraining the low-level controller required approximately 14 hours and utilized around 9 billion simulation timesteps, whereas training the high-level controller along with fine-tuning the low-level policy took approximately 15 hours, covering about 1 billion simulation timesteps.

\begin{table*}[!t]
\centering
\small
\setlength{\tabcolsep}{3.5pt}
\caption{
\ours and the baseline's average tracking error across the four metrics.}
    
\begin{tabular}{l|ccccc}
\toprule
\makecell{Approach\\\textit{Units}}    & \makecell{Pos. mean err. \\$[m] \downarrow$} & \makecell{Ori. mean err.\\$[rad] \downarrow$} & \makecell{Planar vel. mean err.\\$[m/s] \downarrow$} & \makecell{Ang vel. mean err.\\$[rad/s] \downarrow$} & Redundancy \\ 
\midrule
\textbf{\rev{PAKE}} & \rev{\ddpmbf{0.0339}{0.0251}} & \rev{\ddpmbf{0.0481}{0.0720}} & \rev{\ddpmbf{0.1250}{0.0769}} & \rev{\ddpmbf{0.0477}{0.0422}} & \textcolor{ggreen}{\Checkmark} \\
\rev{PAKE-S} & \rev{\ddpm{0.0363}{0.0312}} & \rev{\ddpm{0.0635}{0.1140}} & \rev{\ddpm{0.1266}{0.0885}} & \rev{\ddpm{0.0555}{0.0624}} & \rev{\textcolor{ggreen}{\Checkmark}} \\
\rev{PAKE-S w/o HLC} & \rev{\ddpm{0.0505}{0.0450}} & \rev{\ddpm{0.0715}{0.0925}} & \rev{\ddpm{0.1279}{0.0900}} & \rev{\ddpm{0.0479}{0.0452}} & \rev{\textcolor{gred}{\XSolidBrush}} \\
\midrule
VBC~\cite{liu2024vbc} & \ddpm{0.4431}{0.2781} & \ddpm{1.0816}{0.7765} & \ddpm{0.5070}{0.4795} & \ddpm{0.4499}{0.3793} & \textcolor{gred}{\XSolidBrush} \\
Umi-on-Legs~\cite{ha2024umilegs}       & \ddpm{0.2883}{0.2934} & \ddpm{1.9722}{0.7122} & \textcolor{gred}{\XSolidBrush} & \textcolor{gred}{\XSolidBrush} & \textcolor{gred}{\XSolidBrush} \\
\bottomrule
\end{tabular}
\par\smallskip
\begin{minipage}{\linewidth}\footnotesize
\rev{Redundancy denotes explicit parameterization and learned selection of redundant partial-IK solutions.}
\end{minipage}
\label{tab:baseline}
\end{table*}

\begin{figure*}[!t]
    \centering
    \changedgraphic{\includegraphics[width=\linewidth, trim={0 0 0 0}, clip]{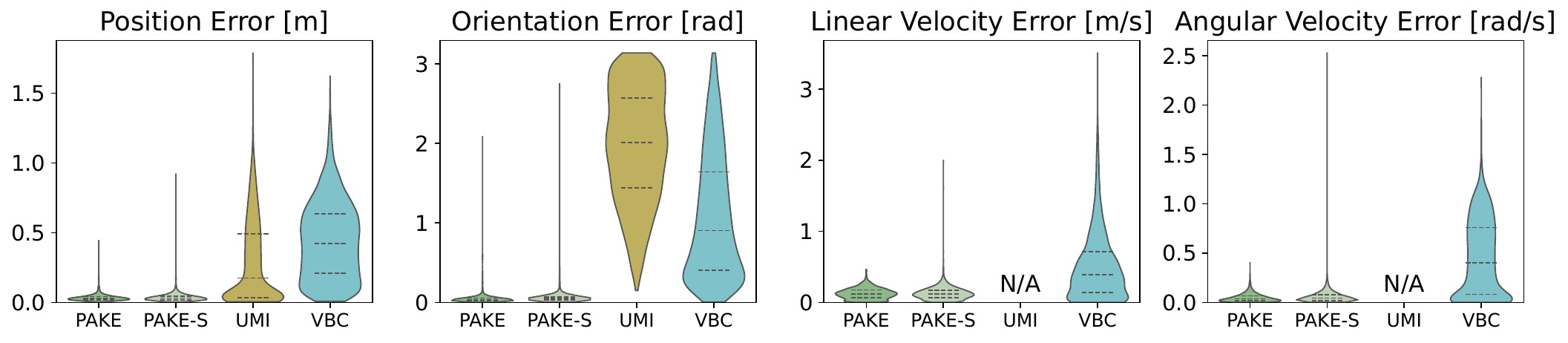}}
    \caption{
    Violin plot distributions comparing tracking errors of different methods across four metrics.
    }
    \label{fig:result_baseline_violin}
\end{figure*}

\begin{figure*}[!t]
    \centering
    \begin{minipage}[t]{0.45\linewidth}
    \centering
    \changedgraphic{\includegraphics[width=\linewidth, trim={0 0 0 0}, clip]{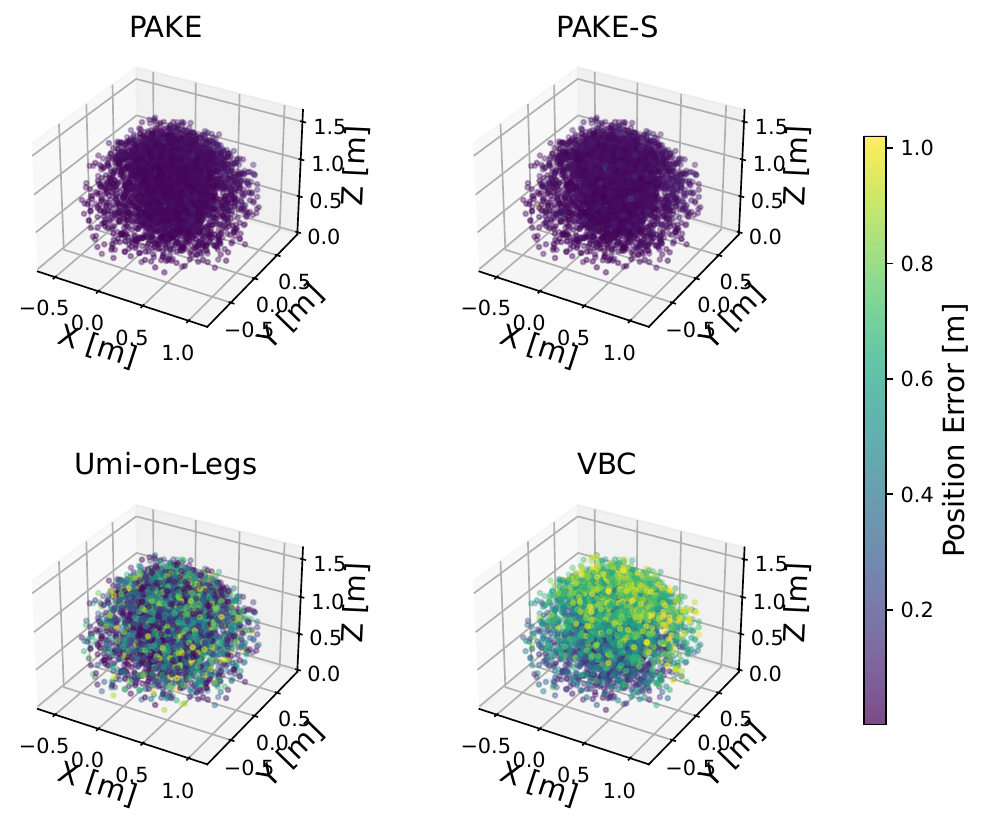}}
    \end{minipage}
    \hspace{0.05\linewidth}
    \begin{minipage}[t]{0.45\linewidth}
    \centering
    \changedgraphic{\includegraphics[width=\linewidth, trim={0 0 0 0}, clip]{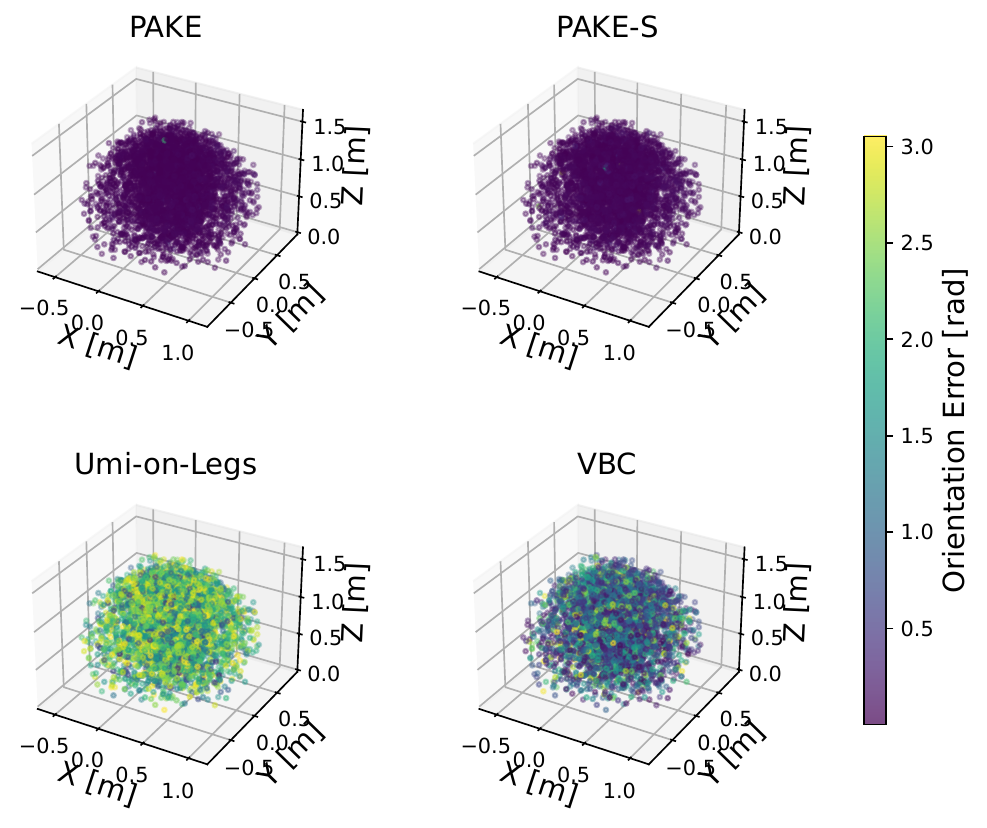}}
    \end{minipage}
    
    \caption{3D visualization of position and orientation tracking effects of \rev{PAKE, PAKE-S, Umi-on-Legs, and VBC}, the color becomes lighter as the tracking error rises. \rev{The color scale saturates at the 99th percentile for display.}}
    \label{fig:3dscatter}
\end{figure*}

\subsection{Performance Comparison}


We \vnew{compare} \vnew{PAKE} \vnew{with} VBC~\cite{liu2024vbc}\vnew{,} Umi-on-Legs~\cite{ha2024umilegs}\vnew{,} \rev{PAKE-S}\vnew{,} \rev{and} \rev{its} \rev{training-without-HLC} \rev{variant}\vnew{.} \vnew{Tracking} \vnew{distributions} \vnew{and} \vnew{feasible} \vnew{workspace} \vnew{complement} \rev{the} \vnew{mean-error} \vnew{comparison}\rev{.}

\begin{itemize}
    \item \textbf{\rev{PAKE-S w/o HLC:}} \rev{The simplified pipeline with the KNF latent fixed to zero throughout training and inference; its LLC is trained for the same 1000 updates from the same initialization.}
    \item \textbf{VBC~\cite{liu2024vbc}:} A commonly used decoupled method in which robotic arm motions are computed via the Jacobian pseudo-inverse method and leg motions are controlled through RL.
    \item \textbf{Umi-on-Legs~\cite{ha2024umilegs}:} \rev{An RL-based whole-body controller trained to track task-specific end-effector trajectories, without independent chassis-velocity commands. For our evaluation, we use its randomized-trajectory variant and perform tracking in the body frame.}
    
\end{itemize}

Both VBC and Umi-on-Legs were trained to convergence in Isaac Gym, as in their respective original papers, using the robot model employed in our baseline evaluations and adjusted parameters.

In our evaluation, each method tracks 4000 randomly sampled 6D end-effector pose targets\rev{, with independent 3D base velocity targets supplied to PAKE and VBC}. Targets are uniformly sampled within predefined bounds using forward kinematics for pose targets (as defined in Section~\ref{subsection:dataset}), and the velocity commands follow the maximum range described in Section~\ref{subsection:command}. Each command remains fixed throughout an episode, ensuring consistent comparisons across methods.

\subsubsection{Mean Tracking Performance Comparison}

Table~\ref{tab:baseline} \vnew{reports} \vnew{mean} \vnew{tracking} \vnew{errors} \vnew{and} \vnew{standard} \vnew{deviations}. \vnew{PAKE} achieves the lowest \vnew{means} across all four metrics. Umi-on-Legs improves position tracking over VBC but has larger orientation errors under the evaluated task settings.

Remarkably, the tracking accuracy achieved by \ours is competitive with concurrent approaches~\cite{jiang2025learning, portela2024wholebodyendeffectorposetracking} (0.022 m, 0.041 rad and 0.026 m, 0.064 rad, respectively). However, these methods are developed under different platform or task assumptions, such as wheeled-legged locomotion or end-effector pose tracking with a stationary base. In contrast, \ours maintains fully legged mobility while coordinating locomotion and manipulation, making it more applicable to loco-manipulation scenarios that require simultaneous leg motion and accurate end-effector tracking.

\subsubsection{Tracking Error Distribution Comparison}


Figure~\ref{fig:result_baseline_violin} shows \vnew{PAKE's} \vnew{concentrated}\vnew{,} \vnew{low-error} \vnew{distributions} \vnew{for} \vnew{pose} \vnew{and} \vnew{torso-velocity} \vnew{tracking}\vnew{.} \vnew{Umi-on-Legs} \vnew{has} \vnew{lower} position \vnew{errors} \vnew{than} \vnew{VBC} \vnew{but} \vnew{larger} \vnew{orientation} \vnew{errors}\vnew{.} \vnew{VBC's} \vnew{Jacobian-based} \vnew{arm} \vnew{control} \vnew{improves} orientation tracking, \vnew{whereas} local minima \vnew{contribute} \vnew{to} \vnew{its} \vnew{broad} \vnew{position-error} \vnew{distribution}. \vnew{PAKE} \vnew{achieves} consistently low \vnew{errors} across all \vnew{four} metrics.

\subsubsection{Feasible Workspace Comparison}



\vnew{Floating-base} \vnew{stability} and arm\vnew{--}\vnew{base} \vnew{coordination} limit precise tracking \vnew{across} the workspace. \vnew{Following}~\cite{fu2022deep}\vnew{,} \vnew{we} define \vnew{feasible} \vnew{workspace} \vnew{using} the convex hull of \vnew{target} \vnew{positions} meeting \vnew{the} \vnew{tracking-error} thresholds\vnew{,} \vnew{excluding} \vnew{its} \vnew{intersection} \vnew{with} \vnew{the} \vnew{torso}. We evaluate position error $<$ 0.1~m, position error $<$ 0.05~m, orientation error $<$ 0.2 rad, and combined position error $<$ 0.1~m \vnew{and} orientation error $<$ 0.2~rad. Table~\ref{tab:workspace} and Figure~\ref{fig:3dscatter} \vnew{show} \vnew{PAKE's} superior feasible workspace coverage.

\begin{table}[!t]
\centering
\scriptsize
\setlength{\tabcolsep}{3.5pt}
\caption{Volume of feasible workspace for different methods at 4 selected thresholds.}

\resizebox{\linewidth}{!}{%
\begin{tabular}{l|cccc}
\toprule
\makecell{Feasible \\ Workspace\( (m^3)\,\uparrow \)}    & Pos Err<0.1 & Pos Err<0.05 & Ori Err<0.2 & \makecell{Pos Err<0.1 \\ Ori Err<0.2} \\ 
\midrule
\rev{\textbf{PAKE}} & \rev{\cc{2.4023}} & \rev{\ccbf{2.3343}} & \rev{\ccbf{2.4072}} & \rev{\cc{2.4023}} \\
\rev{PAKE-S} & \rev{\ccbf{2.4024}} & \rev{\cc{2.2962}} & \rev{\cc{2.4059}} & \rev{\ccbf{2.4024}} \\

\midrule
VBC~\cite{liu2024vbc} & \cc{0.6796} &  \cc{0.3618} &  \cc{1.6113} &  \cc{0.4228} \\
Umi-on-Legs~\cite{ha2024umilegs}  & \cc{2.0628} &  \cc{1.9505} &  \cc{0.0584} &  \cc{0.0174}  \\
\bottomrule
\end{tabular}
}
\label{tab:workspace}
\end{table}

\subsection{\rev{Training and Latent-Input Ablations}}
\begin{revision}
PAKE-S already achieves accurate whole-body tracking. The complete PAKE training further improves tracking accuracy and reduces abrupt changes in the generated references. On the same 4000 targets, the four mean tracking errors decrease by 6.61\%, 24.19\%, 1.27\%, and 14.06\%, respectively (Table~\ref{tab:baseline}). The fraction of arm-reference joint transitions exceeding 3.1415 rad/s decreases from 7.5110\% to 0.6814\%; neither nominal evaluation terminates.

The PAKE-S w/o HLC comparison in Table~\ref{tab:baseline} tests the role of high-level control during training: PAKE-S lowers position, orientation, and planar-velocity errors, while the yaw-rate error is lower without HLC. These results demonstrate the benefit of including HLC in the joint-training procedure.

We hold PAKE fixed and intervene on its KNF latent input. Zero sets the latent to zero; shuffle permutes the learned latent vectors across targets at each step; random uses five fixed per-target draws from PAKE's latent bound. Table~\ref{tab:latent_fixed} reports tracking-error reductions on the same 4000 targets at 3 s, separating the training comparison with PAKE-S from fixed-policy latent interventions.
\end{revision}
\begin{table}[!t]
\centering\small
\setlength{\tabcolsep}{4pt}
\caption{\rev{Tracking-error reductions of PAKE with learned latent inputs relative to each reference condition. Positive values indicate lower errors. PAKE-S evaluates the training contribution; the remaining rows evaluate fixed-policy latent interventions. Random denotes the mean of five draws.}}
\label{tab:latent_fixed}
\begin{revision}
\begin{tabular}{lrrrr}
\toprule
\makecell[l]{Reference\\condition} & $e_p$ [\%] & $e_R$ [\%] & $e_v$ [\%] & $e_\omega$ [\%]\\\midrule
PAKE-S & \textbf{6.61} & \textbf{24.19} & \textbf{1.27} & \textbf{14.06} \\
PAKE $z=0$ & \textbf{3.04} & \textbf{3.31} & \textbf{1.03} & -0.70 \\
PAKE shuffle & \textbf{3.12} & \textbf{4.79} & \textbf{1.18} & \textbf{0.88} \\
PAKE random & \textbf{1.19} & -0.91 & \textbf{0.64} & \textbf{0.76} \\
\bottomrule
\end{tabular}
\par\smallskip
\begin{minipage}{\columnwidth}\footnotesize
Reduction: $100(\bar e_{\mathrm{comparison}}-\bar e_{\mathrm{PAKE}})/\bar e_{\mathrm{comparison}}$. Bold indicates positive reduction, not statistical significance.
\end{minipage}
\end{revision}
\end{table}
\Needspace{4\baselineskip}
\subsection{\rev{Force-Context Latent Swapping}}
\begin{revision}
We evaluate PAKE on a separate set of 200 targets, each repeated across six torso-force contexts: no force, world-frame $\pm x$ and $\pm y$, and the $(+x,+y)$ diagonal. Nonzero forces have magnitude 25 N and act during steps 101--150; base-velocity commands are zero. Native uses each context's learned latent, whereas swap cyclically exchanges the current latent vectors within each same-target group at every step. Table~\ref{tab:force_context} reports endpoint tracking-error reductions at step 300 (3 s), computed from mean errors over all 200 targets and six contexts. Native inputs reduce the four endpoint errors by 4.86\%, 10.23\%, 1.89\%, and 1.93\% relative to swap.
\end{revision}
\begin{table}[!t]
\centering\small
\setlength{\tabcolsep}{4pt}
\caption{\rev{Endpoint tracking-error reductions for Native vs. Swap in the same-target force-context test, relative to Swap errors. Positive values indicate lower errors.}}
\label{tab:force_context}
\begin{revision}
\begin{tabular}{lrrrr}
\toprule
Comparison & $e_p$ [\%] & $e_R$ [\%] & $e_v$ [\%] & $e_\omega$ [\%]\\\midrule
Native vs. Swap & \textbf{4.86} & \textbf{10.23} & \textbf{1.89} & \textbf{1.93} \\
\bottomrule
\end{tabular}
\par\smallskip
\begin{minipage}{\columnwidth}\footnotesize
Reduction: $100(\bar e_{\mathrm{swap}}-\bar e_{\mathrm{native}})/\bar e_{\mathrm{swap}}$. Bold indicates positive reduction, not statistical significance.
\end{minipage}
\end{revision}
\end{table}

\subsection{Hardware Experiments}
\rev{For hardware deployment, we use a policy trained with the PAKE-S pipeline.} We evaluate this PAKE-S policy on a real-world task set comprising 24 episodes (Table~\ref{tab:hardware_summary}). The task set includes \textit{cart\_pull}, \textit{hanger}, \textit{sweep\_broom}, \textit{plug\_in} (Figure~\ref{fig:teaser}), as well as four pick-and-place tasks involving trash disposal, hammer storage, toy pickup, and foam transport. All episodes are evaluated using motion-capture-based end-effector tracking and base-velocity tracking. Across the full task set, PAKE-S achieves a mean end-effector position error of 0.0449~m, a mean orientation error of 0.1437~rad, and a mean planar base-velocity error of 0.0961~m/s.

These results \vnew{demonstrate} stable whole-body \vnew{tracking} across varied object interactions, motion patterns, and mobile manipulation \vnew{tasks}.

\begin{table}[!t]
\centering
\scriptsize
\setlength{\tabcolsep}{3.5pt}
\caption{Aggregate hardware tracking performance of PAKE-S across 24 episodes of 8 tasks. Values are means $\pm$ standard deviations over per-timestep error samples.}
\resizebox{\linewidth}{!}{%
\begin{tabular}{lccccc}
\toprule
Task group & Pos. mean err. & Ori. mean err. & Planar vel. mean err. & Ang vel. mean err. \\
 & [m] $\downarrow$ & [rad] $\downarrow$ & [m/s] $\downarrow$ & [rad/s] $\downarrow$ \\
\midrule
All & \ddpm{0.0449}{0.0100} & \ddpm{0.1437}{0.0164} & \ddpm{0.0961}{0.0355} & \ddpm{0.0145}{0.0123} \\
\midrule
cart\_pull & \ddpm{0.0618}{0.0079} & \ddpm{0.1734}{0.0149} & \ddpm{0.1422}{0.0295} & \ddpm{0.0276}{0.0085} \\
hanger & \ddpm{0.0366}{0.0014} & \ddpm{0.1318}{0.0046} & \ddpm{0.0723}{0.0279} & \ddpm{0.0160}{0.0047} \\
plug\_in & \ddpm{0.0503}{0.0034} & \ddpm{0.1492}{0.0099} & \ddpm{0.0461}{0.0099} & \ddpm{0.0053}{0.0055} \\
sweep\_broom & \ddpm{0.0406}{0.0006} & \ddpm{0.1342}{0.0004} & \ddpm{0.1245}{0.0253} & \ddpm{0.0049}{0.0048} \\
trash disposal & \ddpm{0.0460}{0.0083} & \ddpm{0.1380}{0.0205} & \ddpm{0.1015}{0.0240} & \ddpm{0.0028}{0.0118} \\
hammer storage & \ddpm{0.0433}{0.0054} & \ddpm{0.1508}{0.0146} & \ddpm{0.1028}{0.0135} & \ddpm{0.0229}{0.0171} \\
toy pickup & \ddpm{0.0350}{0.0005} & \ddpm{0.1315}{0.0036} & \ddpm{0.1038}{0.0376} & \ddpm{0.0161}{0.0135} \\
foam transport & \ddpm{0.0452}{0.0150} & \ddpm{0.1407}{0.0032} & \ddpm{0.0753}{0.0115} & \ddpm{0.0205}{0.0092} \\
\bottomrule
\end{tabular}
}
\label{tab:hardware_summary}
\end{table}


\section{Conclusion and future works}

\vnew{We} \vnew{presented} \vnew{PAKE}\rev{,} a \vnew{whole-body} \vnew{loco-manipulation} framework \vnew{that} \rev{explicitly} \vnew{represents} \rev{redundant} \rev{partial-IK} \rev{solutions} and \rev{learns} \rev{to} \rev{select} \vnew{them} \vnew{for} \vnew{accurate} end-effector pose \vnew{and} \vnew{chassis-velocity} tracking. \vnew{Partial} \vnew{torso}\vnew{--}\vnew{arm} reference generation and low-level motion imitation \vnew{preserve} \vnew{omni-directional} \vnew{legged} \vnew{mobility} \vnew{while} \vnew{expanding} \vnew{the} \vnew{manipulator's} \vnew{workspace}. \vnew{Simulation} \vnew{and} \vnew{hardware} \vnew{experiments} \vnew{demonstrate} \vnew{accurate}, \vnew{coordinated} tracking \vnew{across} \vnew{diverse} \vnew{tasks}. The framework \vnew{provides} a \vnew{foundation} \vnew{for} \vnew{more} \vnew{advanced} whole-body control and \vnew{downstream} \vnew{manipulation} applications.



\FloatBarrier
{
\bibliographystyle{IEEEtran}
\bibliography{main}

@IEEEtranBSTCTL{PAKE:IEEEreferenceStyle,
  CTLuse_forced_etal = "yes",
  CTLmax_names_forced_etal = "6",
  CTLnames_show_etal = "1"
}

@ARTICLE{jung2026Learningdynamic,
  author={Jung, Moonkyu and Lee, Jiseong and He, Zhengmao and Youm, Donghoon and Mun, Juhyeok and Kim, HyeongJun and Oh, Hyunsik and Choi, Donghyuk and Hur, Jungwoo and Song, Jie and Hwangbo, Jemin},
  journal={IEEE Robotics and Automation Letters}, 
  title={Learning Dynamic Pick-and-Place for a Legged Manipulator}, 
  year={2026},
  volume={11},
  number={6},
  pages={7652-7659},
  doi={10.1109/LRA.2026.3688092}}

@inproceedings{
cheng2023parkour,
title={Extreme Parkour with Legged Robots},
author={Xuxin Cheng and Kexin Shi and Ananye Agarwal and Deepak Pathak},
booktitle={Towards Generalist Robots: Learning Paradigms for Scalable Skill Acquisition @ CoRL2023},
year={2023}
}

@inproceedings{zhuang2023robot,
  author    = {Zhuang, Ziwen and Fu, Zipeng and Wang, Jianren and Atkeson, Christopher and Schwertfeger, Sören and Finn, Chelsea and Zhao, Hang},
  title     = {Robot Parkour Learning},
  booktitle = {Conference on Robot Learning ({CoRL})},
  year      = {2023},
}

@INPROCEEDINGS{arm2024pedipulate,
  author={Arm, Philip and Mittal, Mayank and Kolvenbach, Hendrik and Hutter, Marco},
  booktitle={2024 IEEE International Conference on Robotics and Automation (ICRA)}, 
  title={Pedipulate: Enabling Manipulation Skills using a Quadruped Robot’s Leg}, 
  year={2024},
  volume={},
  number={},
  pages={5717-5723},
  doi={10.1109/ICRA57147.2024.10611307}}

@article{lee2020learning,
  title={Learning quadrupedal locomotion over challenging terrain},
  author={Lee, Joonho and Hwangbo, Jemin and Wellhausen, Lorenz and Koltun, Vladlen and Hutter, Marco},
  journal={Science robotics},
  volume={5},
  number={47},
  pages={eabc5986},
  year={2020},
  publisher={American Association for the Advancement of Science}
}

@article{hwangbo2019learning,
  title={Learning agile and dynamic motor skills for legged robots},
  author={Hwangbo, Jemin and Lee, Joonho and Dosovitskiy, Alexey and Bellicoso, Dario and Tsounis, Vassilios and Koltun, Vladlen and Hutter, Marco},
  journal={Science Robotics},
  volume={4},
  number={26},
  pages={eaau5872},
  year={2019},
  publisher={American Association for the Advancement of Science}
}

@article{choi2023deformable,
  title={Learning quadrupedal locomotion on deformable terrain},
  author={Choi, Suyoung and Ji, Gwanghyeon and Park, Jeongsoo and Kim, Hyeongjun and Mun, Juhyeok and Lee, Jeong Hyun and Hwangbo, Jemin},
  journal={Science Robotics},
  volume={8},
  number={74},
  pages={eade2256},
  year={2023},
  publisher={American Association for the Advancement of Science}
}

@INPROCEEDINGS{ji2023Hierarchical,
  author={Ji, Yandong and Li, Zhongyu and Sun, Yinan and Peng, Xue Bin and Levine, Sergey and Berseth, Glen and Sreenath, Koushil},
  booktitle={2022 IEEE/RSJ International Conference on Intelligent Robots and Systems (IROS)}, 
  title={Hierarchical Reinforcement Learning for Precise Soccer Shooting Skills using a Quadrupedal Robot}, 
  year={2022},
  volume={},
  number={},
  pages={1479-1486},
  doi={10.1109/IROS47612.2022.9981984}}

@inproceedings{fu2022deep,
  author    = {Fu, Zipeng and Cheng, Xuxin and Pathak, Deepak},
  title     = {Deep Whole-Body Control: Learning a Unified Policy for Manipulation and Locomotion},
  booktitle = {Conference on Robot Learning ({CoRL})},
  year      = {2022},
}

@article{liu2024vbc,
title={Visual Whole-Body Control for Legged Loco-Manipulation},
author={Liu, Minghuan and Chen, Zixuan and Cheng, Xuxin and Ji, Yandong and Qiu, Rizhao and Yang, Ruihan and Wang, Xiaolong},
journal={The 8th Conference on Robot Learning},
year={2024}
}

@inproceedings{
ha2024umilegs,
title={{UMI}-on-Legs: Making Manipulation Policies Mobile with Manipulation-Centric Whole-body Controllers},
author={Huy Ha and Yihuai Gao and Zipeng Fu and Jie Tan and Shuran Song},
booktitle={8th Annual Conference on Robot Learning},
year={2024},
url={https://openreview.net/forum?id=3i7j8ZPnbm}
}

@INPROCEEDINGS{portela2023learningforce,
  author={Portela, Tifanny and Margolis, Gabriel B. and Ji, Yandong and Agrawal, Pulkit},
  booktitle={2024 IEEE International Conference on Robotics and Automation (ICRA)}, 
  title={Learning Force Control for Legged Manipulation}, 
  year={2024},
  volume={},
  number={},
  pages={15366-15372},
  doi={10.1109/ICRA57147.2024.10611066}}

@INPROCEEDINGS{he2024legmanip,
  author={He, Zhengmao and Lei, Kun and Ze, Yanjie and Sreenath, Koushil and Li, Zhongyu and Xu, Huazhe},
  booktitle={2024 IEEE/RSJ International Conference on Intelligent Robots and Systems (IROS)}, 
  title={Learning Visual Quadrupedal Loco-Manipulation from Demonstrations}, 
  year={2024},
  volume={},
  number={},
  pages={9102-9109},
  doi={10.1109/IROS58592.2024.10802742}}

@article{ma2022combining,
  title={Combining learning-based locomotion policy with model-based manipulation for legged mobile manipulators},
  author={Ma, Yuntao and Farshidian, Farbod and Miki, Takahiro and Lee, Joonho and Hutter, Marco},
  journal={IEEE Robotics and Automation Letters},
  volume={7},
  number={2},
  pages={2377--2384},
  year={2022},
  publisher={IEEE}
}

@article{ferrolho2023roloma,
  title={Roloma: Robust loco-manipulation for quadruped robots with arms},
  author={Ferrolho, Henrique and Ivan, Vladimir and Merkt, Wolfgang and Havoutis, Ioannis and Vijayakumar, Sethu},
  journal={Autonomous Robots},
  volume={47},
  number={8},
  pages={1463--1481},
  year={2023},
  publisher={Springer}
}

@INPROCEEDINGS{belliscoso2019alma,
  author={Bellicoso, C. Dario and Krämer, Koen and Stäuble, Markus and Sako, Dhionis and Jenelten, Fabian and Bjelonic, Marko and Hutter, Marco},
  booktitle={2019 International Conference on Robotics and Automation (ICRA)}, 
  title={ALMA - Articulated Locomotion and Manipulation for a Torque-Controllable Robot}, 
  year={2019},
  volume={},
  number={},
  pages={8477-8483},
  doi={10.1109/ICRA.2019.8794273}}

@ARTICLE{sleiman2021mpc,
  author={Sleiman, Jean-Pierre and Farshidian, Farbod and Minniti, Maria Vittoria and Hutter, Marco},
  journal={IEEE Robotics and Automation Letters}, 
  title={A Unified MPC Framework for Whole-Body Dynamic Locomotion and Manipulation}, 
  year={2021},
  volume={6},
  number={3},
  pages={4688-4695},
  doi={10.1109/LRA.2021.3068908}}

@inproceedings{
makoviychuk2021isaac,
title={Isaac Gym: High Performance {GPU} Based Physics Simulation For Robot Learning},
author={Viktor Makoviychuk and Lukasz Wawrzyniak and Yunrong Guo and Michelle Lu and Kier Storey and Miles Macklin and David Hoeller and Nikita Rudin and Arthur Allshire and Ankur Handa and Gavriel State},
booktitle={Thirty-fifth Conference on Neural Information Processing Systems Datasets and Benchmarks Track (Round 2)},
year={2021}
}

@ARTICLE{jiang2025learning,
  author={Jiang, Kaiwen and Fu, Zhen and Guo, Junde and Zhang, Wei and Chen, Hua},
  journal={IEEE Robotics and Automation Letters}, 
  title={Learning Whole-Body Loco-Manipulation for Omni-Directional Task Space Pose Tracking With a Wheeled-Quadrupedal-Manipulator}, 
  year={2025},
  volume={10},
  number={2},
  pages={1481-1488},
  doi={10.1109/LRA.2024.3519856}}

@misc{portela2024wholebodyendeffectorposetracking,
      title={Whole-body end-effector pose tracking}, 
      author={Tifanny Portela and Andrei Cramariuc and Mayank Mittal and Marco Hutter},
      year={2024},
      eprint={2409.16048},
      archivePrefix={arXiv},
      primaryClass={cs.RO},
      url={https://arxiv.org/abs/2409.16048}, 
}

@ARTICLE{pan2024roboduetwholebodyleggedlocomanipulation,
  author={Pan, Guoping and Ben, Qingwei and Yuan, Zhecheng and Jiang, Guangqi and Ji, Yandong and Li, Shoujie and Pang, Jiangmiao and Liu, Houde and Xu, Huazhe},
  journal={IEEE Robotics and Automation Letters}, 
  title={RoboDuet: Learning a Cooperative Policy for Whole-Body Legged Loco-Manipulation}, 
  year={2025},
  volume={10},
  number={5},
  pages={4564-4571},
  doi={10.1109/LRA.2025.3551230}}

@inproceedings{kingma2018glow,
 author = {Kingma, Durk P and Dhariwal, Prafulla},
 booktitle = {Advances in Neural Information Processing Systems},
 editor = {S. Bengio and H. Wallach and H. Larochelle and K. Grauman and N. Cesa-Bianchi and R. Garnett},
 pages = {},
 publisher = {Curran Associates, Inc.},
 title = {Glow: Generative Flow with Invertible 1x1 Convolutions},
 url = {https://proceedings.neurips.cc/paper_files/paper/2018/file/d139db6a236200b21cc7f752979132d0-Paper.pdf},
 volume = {31},
 year = {2018}
}

@inproceedings{kim2020softflow,
 author = {Kim, Hyeongju and Lee, Hyeonseung and Kang, Woo Hyun and Lee, Joun Yeop and Kim, Nam Soo},
 booktitle = {Advances in Neural Information Processing Systems},
 editor = {H. Larochelle and M. Ranzato and R. Hadsell and M.F. Balcan and H. Lin},
 pages = {16388--16397},
 publisher = {Curran Associates, Inc.},
 title = {SoftFlow: Probabilistic Framework for Normalizing Flow on Manifolds},
 url = {https://proceedings.neurips.cc/paper_files/paper/2020/file/bdbca288fee7f92f2bfa9f7012727740-Paper.pdf},
 volume = {33},
 year = {2020}
}

@ARTICLE{ames2022ikflow,
  author={Ames, Barrett and Morgan, Jeremy and Konidaris, George},
  journal={IEEE Robotics and Automation Letters}, 
  title={IKFlow: Generating Diverse Inverse Kinematics Solutions}, 
  year={2022},
  volume={7},
  number={3},
  pages={7177-7184},
  doi={10.1109/LRA.2022.3181374}}

@INPROCEEDINGS{Brakel2022asymmetric,
  author={Brakel, Philémon and Bohez, Steven and Hasenclever, Leonard and Heess, Nicolas and Bousmalis, Konstantinos},
  booktitle={2022 IEEE/RSJ International Conference on Intelligent Robots and Systems (IROS)}, 
  title={Learning Coordinated Terrain-Adaptive Locomotion by Imitating a Centroidal Dynamics Planner}, 
  year={2022},
  volume={},
  number={},
  pages={10335-10342},
  doi={10.1109/IROS47612.2022.9981648}}

@ARTICLE{ji2022concurrent,
  author={Ji, Gwanghyeon and Mun, Juhyeok and Kim, Hyeongjun and Hwangbo, Jemin},
  journal={IEEE Robotics and Automation Letters}, 
  title={Concurrent Training of a Control Policy and a State Estimator for Dynamic and Robust Legged Locomotion}, 
  year={2022},
  volume={7},
  number={2},
  pages={4630-4637},
  doi={10.1109/LRA.2022.3151396}}

@article{raisim,
  title={Per-contact iteration method for solving contact dynamics},
  author={Hwangbo, Jemin and Lee, Joonho and Hutter, Marco},
  journal={IEEE Robotics and Automation Letters},
  url="www.raisim.com",
  volume={3},
  number={2},
  pages={895--902},
  year={2018},
  publisher={IEEE}
}

@inproceedings{pytorch,
 author = {Paszke, Adam and Gross, Sam and Massa, Francisco and Lerer, Adam and Bradbury, James and Chanan, Gregory and Killeen, Trevor and Lin, Zeming and Gimelshein, Natalia and Antiga, Luca and Desmaison, Alban and Kopf, Andreas and Yang, Edward and DeVito, Zachary and Raison, Martin and Tejani, Alykhan and Chilamkurthy, Sasank and Steiner, Benoit and Fang, Lu and Bai, Junjie and Chintala, Soumith},
 booktitle = {Advances in Neural Information Processing Systems},
 editor = {H. Wallach and H. Larochelle and A. Beygelzimer and F. d\textquotesingle Alch\'{e}-Buc and E. Fox and R. Garnett},
 pages = {},
 publisher = {Curran Associates, Inc.},
 title = {PyTorch: An Imperative Style, High-Performance Deep Learning Library},
 url = {https://proceedings.neurips.cc/paper_files/paper/2019/file/bdbca288fee7f92f2bfa9f7012727740-Paper.pdf},
 volume = {32},
 year = {2019}
}

@article{hwangbo2025raibo2,
  title={A quadruped robot designed to complete a marathon on a single battery charge},
  author={Lee, Choongin and Youm, Donghoon and Park, Jeongsoo and Lee, Jeonghyun and Choi, Suyoung and Ji, Gwanghyeon and Mun, Juhyeok and Jung, Moonkyu and Choi, Hyungho Chris and Kim, Hyeongjun and others},
  journal={Nature},
  pages={1--8},
  year={2026},
  publisher={Nature Publishing Group UK London}
}

@INPROCEEDINGS{bell2016perception,
  author={Dario Bellicoso, C. and Gehring, Christian and Hwangbo, Jemin and Fankhauser, Péter and Hutter, Marco},
  booktitle={2016 IEEE-RAS 16th International Conference on Humanoid Robots (Humanoids)}, 
  title={Perception-less terrain adaptation through whole body control and hierarchical optimization}, 
  year={2016},
  volume={},
  number={},
  pages={558-564},
  doi={10.1109/HUMANOIDS.2016.7803330}}

@INPROCEEDINGS{Carlo2018cheetah,
  author={Di Carlo, Jared and Wensing, Patrick M. and Katz, Benjamin and Bledt, Gerardo and Kim, Sangbae},
  booktitle={2018 IEEE/RSJ International Conference on Intelligent Robots and Systems (IROS)}, 
  title={Dynamic Locomotion in the MIT Cheetah 3 Through Convex Model-Predictive Control}, 
  year={2018},
  volume={},
  number={},
  pages={1-9},
  doi={10.1109/IROS.2018.8594448}}

@article{kim2019wbc,
  title={Highly dynamic quadruped locomotion via whole-body impulse control and model predictive control},
  author={Kim, Donghyun and Di Carlo, Jared and Katz, Benjamin and Bledt, Gerardo and Kim, Sangbae},
  journal={arXiv preprint arXiv:1909.06586},
  year={2019}
}

@ARTICLE{Ruben2023nonlinear,
  author={Grandia, Ruben and Jenelten, Fabian and Yang, Shaohui and Farshidian, Farbod and Hutter, Marco},
  journal={IEEE Transactions on Robotics}, 
  title={Perceptive Locomotion Through Nonlinear Model-Predictive Control}, 
  year={2023},
  volume={39},
  number={5},
  pages={3402-3421},
  doi={10.1109/TRO.2023.3275384}}

@INPROCEEDINGS{ma2023fall,
  author={Ma, Yuntao and Farshidian, Farbod and Hutter, Marco},
  booktitle={2023 IEEE International Conference on Robotics and Automation (ICRA)}, 
  title={Learning Arm-Assisted Fall Damage Reduction and Recovery for Legged Mobile Manipulators}, 
  year={2023},
  volume={},
  number={},
  pages={12149-12155},
  doi={10.1109/ICRA48891.2023.10160582}}

@article{
yang2020multiexpert,
author = {Chuanyu Yang  and Kai Yuan  and Qiuguo Zhu  and Wanming Yu  and Zhibin Li },
title = {Multi-expert learning of adaptive legged locomotion},
journal = {Science Robotics},
volume = {5},
number = {49},
pages = {eabb2174},
year = {2020},
doi = {10.1126/scirobotics.abb2174},
URL = {https://www.science.org/doi/abs/10.1126/scirobotics.abb2174},
eprint = {https://www.science.org/doi/pdf/10.1126/scirobotics.abb2174},
}

@inproceedings{
margolis2022walkwaystuningrobot,
title={Walk These Ways: Tuning Robot Control for Generalization with Multiplicity of Behavior},
author={Gabriel B. Margolis and Pulkit Agrawal},
booktitle={6th Annual Conference on Robot Learning},
year={2022},
url={https://openreview.net/forum?id=52c5e73SlS2}
}

@article{
hoeller2023anymal,
author = {David Hoeller  and Nikita Rudin  and Dhionis Sako  and Marco Hutter },
title = {ANYmal parkour: Learning agile navigation for quadrupedal robots},
journal = {Science Robotics},
volume = {9},
number = {88},
pages = {eadi7566},
year = {2024},
doi = {10.1126/scirobotics.adi7566},
URL = {https://www.science.org/doi/abs/10.1126/scirobotics.adi7566},
eprint = {https://www.science.org/doi/pdf/10.1126/scirobotics.adi7566}}

@article{zhang2025RobustDexGrasp,
  title={{RobustDexGrasp}: Robust Dexterous Grasping of General Objects from Single-view Perception},
  author={Zhang, Hui and Wu, Zijian and Huang, Linyi and Christen, Sammy and Song, Jie},
  journal={arXiv preprint arXiv:2504.05287},
  year={2025}
}

@article{huang2024fungrasp,
  title={{FunGrasp}: Functional Grasping for Diverse Dexterous Hands},
  author={Huang, Linyi and Zhang, Hui and Wu, Zijian and Christen, Sammy and Song, Jie},
  journal=RAL,
  year={2025}
}
}


\begin{acronym}
\acro{HP}{high-pass}
\acro{LP}{low-pass}
\end{acronym}

\end{document}